\documentclass[sigconf,screen]{acmart}
\renewcommand\footnotetextcopyrightpermission[1]{}

\usepackage{float}
\usepackage{multirow}
\usepackage{diagbox}
\usepackage{balance}

\usepackage[linesnumbered,titlenumbered,ruled,vlined,commentsnumbered,noend]{algorithm2e}

\usepackage[T1]{fontenc}

\usepackage{xcolor}
\usepackage{epsfig}
\usepackage{pifont} 

\usepackage{enumitem}
\usepackage{moresize}
\usepackage{epstopdf}
\usepackage{array}

\usepackage{bm}
\usepackage{dsfont}
\usepackage{indentfirst}

\usepackage{amssymb}
\usepackage{sidecap}
\usepackage{colortbl}

\usepackage[font=small,labelfont=bf]{caption}

\usepackage{wrapfig}
\usepackage{subfigure}
\usepackage[hang,flushmargin]{footmisc} 

\def\argmax{\operatornamewithlimits{arg\,max}}

\definecolor{royalblue}{RGB}{65,105,225} %
\usepackage{hyperref}

\usepackage{xspace}

\makeatletter
\DeclareRobustCommand\onedot{\futurelet\@let@token\@onedot}
\def\@onedot{\ifx\@let@token.\else.\null\fi\xspace}
\def\eg{\emph{e.g}\onedot} 
\def\ie{\emph{i.e}\onedot}

\makeatother

\definecolor{darkgreen}{RGB}{63, 143, 49}
\definecolor{darkred}{RGB}{181, 56, 56}

\definecolor{darkorange}{rgb}{1.0, 0.55, 0.0}
\definecolor{lincolngreen}{rgb}{0.11, 0.35, 0.02}
\definecolor{cornflowerblue}{rgb}{0.39, 0.58, 0.93} 

\definecolor{cobalt}{rgb}{0.0, 0.28, 0.67}

\begin{document}

\title{Masked Faces with Faced Masks}

\author{
Jiayi Zhu$^1$, \ Qing Guo$^{2}$, \ Felix Juefei-Xu$^3$, \ Yihao Huang$^1$, \ Yang Liu$^{2}$, \ Geguang Pu$^1$\\ }
\affiliation{\institution{
$^1$ East China Normal University\country{China}\\
$^2$ Nanyang Technological University\country{Singapore}\\
$^3$ Alibaba Group\country{USA}\\}
}


\begin{abstract}
Modern face recognition systems (FRS) still fall short when the subjects are wearing facial masks, a common theme in the age of respiratory pandemics. An intuitive partial remedy is to add a mask detector to flag any masked faces so that the FRS can act accordingly for those low-confidence masked faces.
In this work, we set out to investigate the potential vulnerability of such FRS equipped with a mask detector, on large-scale masked faces, which might trigger a serious risk, \eg, letting a suspect evade the FRS where both facial identity and mask are undetected.
As existing face recognizers and mask detectors have high performance in their respective tasks, \textit{it is significantly challenging to simultaneously fool them and preserve the transferability of the attack.}
We formulate the new task as the generation of realistic~\&~adversarial-faced mask and make three main contributions: \textit{First}, we study the naive \textit{Delanunay-based masking method (DM)} to simulate the process of wearing a faced mask that is cropped from a template image, which reveals the main challenges of this new task. \textit{Second}, we further equip the DM with the adversarial noise attack and propose the \textit{adversarial noise Delaunay-based masking method (AdvNoise-DM)} that can fool the face recognition and mask detection effectively but make the face less natural. \textit{Third}, we propose the \textit{adversarial filtering Delaunay-based masking method} denoted as $\text{MF}^2\text{M}$ by employing the adversarial filtering for AdvNoise-DM and obtain more natural faces. 
With the above efforts, the final version not only leads to significant performance deterioration of the state-of-the-art (SOTA) deep learning-based FRS, but also remains undetected by the SOTA facial mask detector, thus successfully fooling both systems at the same time. 
We conduct extensive white-box and black-box experiments on three FRS and a facial mask detector.
We utilize the datasets in MegaFace Challenge 1 and evaluate on dimensions of face recognition, face verification and mask detection, comparing with the solid-colored masking method and seven SOTA adversarial attacks.
Moreover, we also set up the physical experiments by printing the adversarial faces in the real world and re-capturing them to fool the face recognition and mask detector, which demonstrates the high generalizability of our method.
Overall, the proposed method, for the first attempt, unveils the vulnerability of the FRS when dealing with masked faces wearing faced masks.
\end{abstract}


\begin{teaserfigure}
  \includegraphics[width=\textwidth]{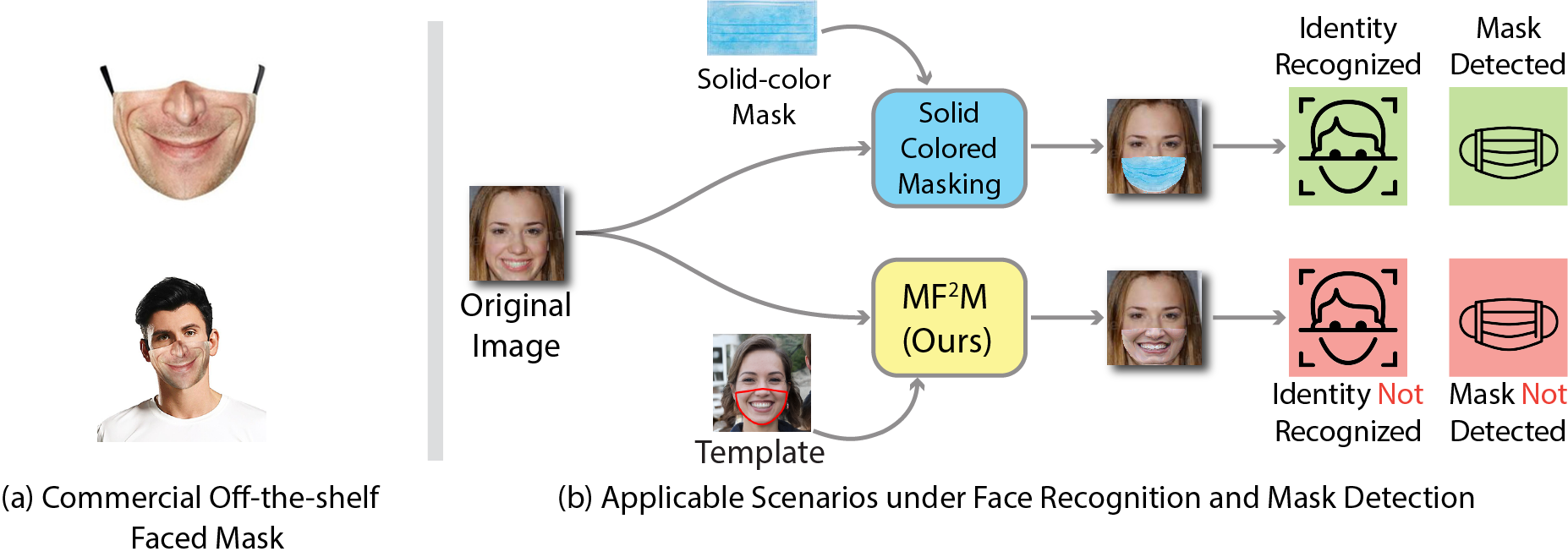}
  \vspace{-20pt}
  \caption{(a) shows a commercial off-the-shelf (COTS) faced mask and the look of being worn. 
    (b) represents the security problem we are exploring.
    Faces wearing solid-color masks will be recognized as their original identities in most cases and easily detected by a mask detector (the upper arrow).
    The $\text{MF}^2\text{M}$ we proposed extracts the face information in the area surrounded by the red line of the template image to obtain a ``faced mask'', which can simultaneously deceive face recognizers and evade mask detection (the lower arrow).}
  \label{fig:fig1}
  \vspace{10pt}
\end{teaserfigure}

\maketitle

\section{Introduction}

Currently, under the severe international situation and environment (\ie, COVID-19 pandemic), people are mandatorily required to wear facial masks in public, especially in crowded places like airports.
This situation poses a huge challenge for face recognition systems (FRS).
Although existing face recognition models (\eg, SphereFace \cite{liu2017sphereface}, CosFace \cite{wang2018cosface}, ArcFace \cite{deng2018arcface}) have high-performance on identity recognition tasks, these models are only available to faces in good imagery conditions.
When faces are heavily obscured (\eg, wearing facial masks), even the state-of-the-art (SOTA) FRS do not perform satisfactorily since the information of the masked area is lost.
Recently, the National Institute of Standards and Technology (NIST) published a specific study to confirm that the accuracy of FRS drops sharply targeting masked faces \cite{ngan2020ongoing}.

An indirect way to solve this problem is to do mask detection.
Once a facial mask is detected, the inspector can be made aware of the inaccuracy of the current face recognition result and respond accordingly.
However, there are various styles of commercial off-the-shelf (COTS) facial masks, even some faced masks (\ie, printed with the lower half of faces from celebrities) as shown in Fig.~\ref{fig:fig1}(a).
Such COTS faced masks cause great confusion to existing mask detectors as these detectors only consider solid-colored masks (\eg, common surgical masks) during training.
Therefore, existing mask detectors can not deal with masks with special textures and complex patterns.
It is worrying that potential offenders may wear such COTS faced masks and even do special treatment to viciously hide their identities and avoid mask detection at the same time.
To explore this security problem, we simulate the process of manufacturing masks with face patterns and propose ``faced mask'' approaches.
Our approaches attack both the FRS and the mask detector, exposing their weaknesses under this specific multitasking attack.

In this paper, we propose an adversarial filtering Delaunay-based masking method, denotes as {\textbf{Masked Faces with Faced Masks} ($\text{MF}^2\text{M}$)}, to stealthily generate masks with face patterns. The perpetrating faced masks not only significantly reduce the accuracy of two SOTA deep learning-based FRS but also drop the accuracy of a SOTA mask detector by 83.58\%.
As shown in Fig.~\ref{fig:fig1}(b), faces wearing solid-colored masks will be recognized as their original identities in most cases and easily detected by a mask detector (the upper arrow).
The $\text{MF}^2\text{M}$ we proposed (the lower arrow) can simultaneously deceive face recognizers and evade mask detection.
In particular, we first modify the Delaunay method \cite{delaunaytriangulation} to simulate the process of wearing masks and propose a Delaunay-based masking method.
We replace the lower face of the input image (\ie, the original image in Fig.~\ref{fig:fig1}(b)) with the lower face of the desired face image (\ie, the area surrounded by the red line of the template image in Fig.~\ref{fig:fig1}(b)).
Intuitively, adding adversarial noise to the mask can successfully attack both the face recognition and mask detection systems. 
However, our experiments show that this method will cause the pixels to be strongly modified.
To make up for this deficiency and make images look more natural, we further exploit the advantages of filters and propose the novel filtering-based attack method $\text{MF}^2\text{M}$.

Since the FRS mainly takes features of the eye area (also known as the periocular region \cite{juefei2015spartans,juefei2014subspace,fx_dissertation,juefei2016fastfood,juefei2014hallucinating}) into consideration, attacking the FRS through only modifying the lower face is much more difficult than modifying the upper face. 
To our best knowledge, previous methods all attack the FRS by modifying the upper face area \cite{sharif2016accessorize,komkov2021advhat,yin2021adv}.
Our method is the first attempt which only changes the lower face area to attack the FRS. 
Furthermore, our method (\ie, $\text{MF}^2\text{M}$) can both be used as white-box and black-box attacks to the SOTA FRS, which reflects its usability and universality.

The contributions are summarized as follows.
\ding{182} We study the naive Delanunay-based masking method (DM) to simulate the process of wearing a faced mask, which reveals the main challenge that this operation of only replacing the lower face does not strongly interference discriminators.
\ding{183} We further equip the DM with the adversarial noise attack and propose the adversarial noise Delaunay-based masking method (AdvNoise-DM) that can handle the joint-task that fools the face recognition and mask detection effectively.
\ding{184} We propose the adversarial filtering Delaunay-based masking method (denoted as $\text{MF}^2\text{M}$) by employing the adversarial filtering for AdvNoise-DM.
This masking method leads to significant performance deterioration of SOTA deep learning-based face recognizers and mask detector while ensuring the naturalness of the obtained faces.
\ding{185} Our extensive experiments in white-box attack and black-box attack demonstrate the universality and transferability of our proposed $\text{MF}^2\text{M}$.
Then we extend to physical attack and illustrate the robustness of our proposed masking method.
%

\section{Related Work}

\noindent\textbf{{Face recognition.}}
Face recognition can be divided into closed-set recognition and open-set recognition. 
For closed-set recognition, the identities in the test set need to be included in the training datasets. This task is regarded as a multi-class classification problem and is usually solved by a softmax classifier \cite{taigman2014deepface,sun2014deep,parkhi2015deep,cao2018vggface2,wang2017multi}.
%
Currently, most face recognition researches focus on open-set recognition (\ie, identities for testing do not exist in the training datasets).
A series of algorithms were proposed to learn an embedding to represent each identity \cite{liu2017sphereface,wang2018cosface,deng2018arcface,wang2018additive,liu2016large,wen2016discriminative,zhang2017range}.
Those algorithms modify the loss function to maximize inter-class variance and minimize intra-class variance. 
SphereFace \cite{liu2017sphereface} pioneered an angular softmax loss to learn angularly discriminative features. 
CosFace \cite{wang2018cosface} presented a large margin cosine loss to remove radial variations.
%
ArcFace \cite{deng2018arcface} proposed an additive angular margin loss. 
Although these methods show high performance on the face recognition task under good imagery conditions, they pay little attention to obscured faces and are incompetent with masked faces recognition.

To tackle this problem, some researchers added specific modules to adjust existing models and strengthen the recognition performance for masked faces \cite{li2021cropping,montero2021boosting,boutros2021unmasking}.
Recently, Masked Face Recognition Competition (MFR 2021) \cite{boutros2021mfr} was held to promote masked face recognition accuracy. 
Almost all of the participants utilized variations of the ArcFace as their loss and exploited either real or simulated solid-colored masked face images as part of their training datasets. 
These masked face recognition methods showed better robustness regarding faces masked with solid-colored masks.
However, these methods do not solve the problem essentially, not considering masks with face patterns and special textures.

\begin{figure*}
\centering
\includegraphics[width=1\linewidth]{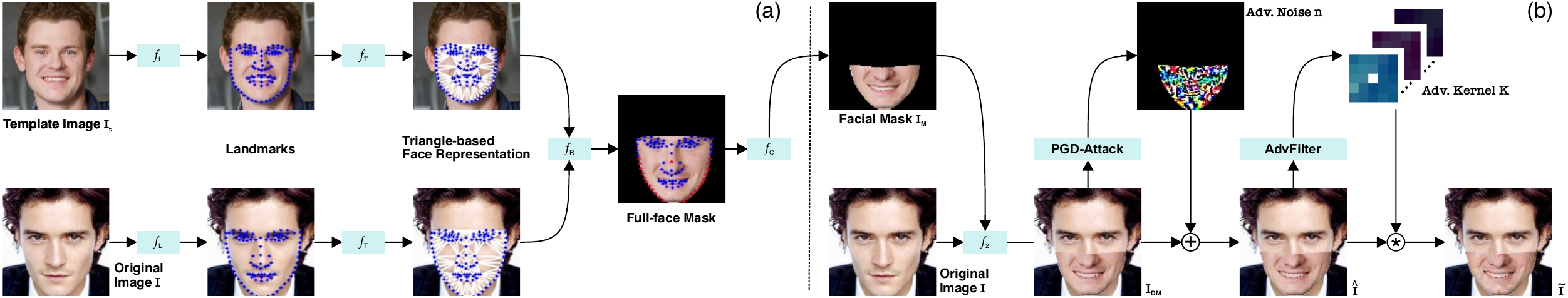}
\vspace{-20pt}
\caption{(a) shows process of generating the faced mask from the original image and the template image. (b) shows the pipeline of the proposed Delaunay-based Masking method, AdvNoise-DM, and $\text{MF}^2\text{M}$.
}
\label{fig:pipeline}
\vspace{-10pt}
\end{figure*}
\vspace{5pt}
\noindent\textbf{{Mask detection.}}
In the age of the outbreak of the COVID-19 pandemic, researchers pay more attention to masked face detection and relative datasets.
The Masked Faces (MAFA) dataset \cite{ge2017detecting} is an early proposed dataset for occluded face detection, which is collected from the Internet, varying in pose angle and occlusion degree. 
%
Recently, Wuhan University has introduced the Real-world Masked Face Recognition Dataset (RMFRD) and the Simulated Masked Face Recognition Dataset (SMFRD) \cite{wang2020masked}. 
These datasets all focus on masks with solid colors and ignore masks with specific textures or complex patterns.
To judge whether there is a mask, some mask detection methods fine-tune face detection models to meet the requirement \cite{batagelj2021correctly,loey2021hybrid,qin2020identifying}. 
Although these methods perform well on common mask detection, they can not deal with special masks with facial textures as they ignore this situation.

\vspace{5pt}
\noindent\textbf{{Adversarial attack.}}
There are a series of adversarial attacks. 
The fast gradient sign method (FGSM) \cite{goodfellow2014explaining} first proposes the additive-perturbation-based attack and the iterative fast gradient sign method (I-FGSM) \cite{kurakin2016adversarial} is an iterative variant of FGSM.
Then the momentum iterative fast gradient sign method (MI-FGSM) \cite{dong2018boosting} introduces the idea of momentum, which helps to stabilize optimization and escape from poor local maxima in the iteration.
Besides, the translation-invariant fast gradient sign method (TI-FGSM) \cite{dong2019evading} and the diverse inputs iterative fast gradient sign method ($\text{DI}^2$-FGSM) \cite{xie2019improving} are both designed for transferability and apply transformations to the input images at each iteration in the attack process.
TI-FGSM utilizes a kernel matrix to simulate the translation of images in different directions while $\text{DI}^2$-FGSM applies random resizing and padding to images with a given probability.
More recently, a series of works focus on natural degradation-based adversarial attacks like adversarial morphing attack against face recognition \cite{wang2020amora}, adversarial relighting attack \cite{gao2021adversarial}, adversarial blur attack \cite{neurips20_abba,guo2021learning}, and adversarial vignetting attack \cite{tian2021ava}.  
These works explore the robustness of deep models by adding natural degradations like motion blur and light variation to the input with different adversarial objective functions. In this work, we actually regard the faced masks as the real-world perturbations.

\section{Methodology}
In order to illustrate the potential hazards of such faced masks, we propose a multi-stage framework to interfere with face recognizers and avoid mask detection simultaneously.
%

\subsection{Delaunay-based Masking and Motivation}
\label{subsec:faceswap}

\begin{figure}
	\centering 
 	\includegraphics[width=0.85\columnwidth]{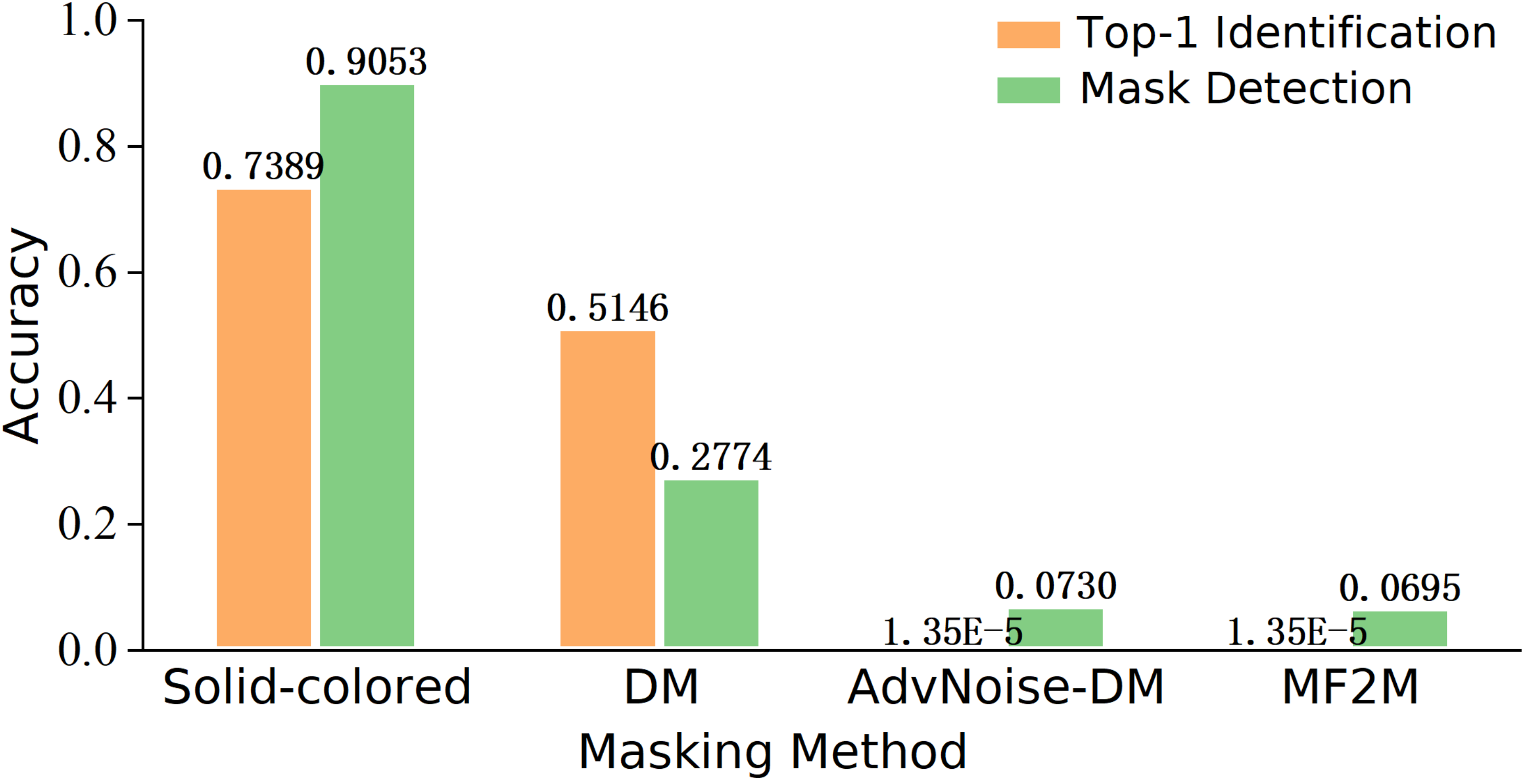}
 	\vspace{-10pt}
	\caption{
    %
    %
    Top-1 identification rates and mask detection rates for solid-colored mask, DM, AdvNoise-DM and $\text{MF}^2\text{M}$ on MegaFace Challenge 1.
    }
	\label{fig:bar}
	\vspace{-10pt}
\end{figure}

%
To simulate the process of wearing faced masks, the most intuitive way is to do an operation similar to face replacing in the specified area. 
We first apply a Delaunay-based masking method (DM).
This method operates on two images, one is an original image $\mathbf{I}\in \mathbb{R}^{H_1\times W_1\times 3}$ on which we want to put the faced mask, the other is a template image $\mathbf{I_t}\in \mathbb{R}^{H_2\times W_2\times 3}$ used to build the faced mask. 
The template image can be unreal and constructed by some DeepFake technique (\eg, StyleGAN \cite{karras2019style}).
We aim to generate a facial mask that contains partial face patterns from $\mathbf{I_t}$ and
looks natural when added to $\mathbf{I}$ with the process
%
\begin{align}
    \mathbf{I_{M}} = f_1(\mathbf{I}, \mathbf{I_t}),
    \label{eq:faceswap_DM}
\end{align}
where $\mathbf{I_M}\in \mathbb{R}^{H_1\times W_1\times 3}$ is the obtained mask (\eg, facial mask in Fig.~\ref{fig:pipeline}).
Specifically, we expand the function $f_1(\cdot)$ as
\begin{align}\label{eq:faceswap-1}
    f_1(\mathbf{I}, \mathbf{I_t}) = f_\text{C}(f_\text{R}(f_\text{T}(f_\text{L}(\mathbf{I}_\mathbf{t})), f_\text{T}(f_\text{L}(\mathbf{I})))),
\end{align}
where $f_\text{L}(\cdot)$ extracts the landmarks of $\mathbf{I}_\mathbf{t}$ and $\mathbf{I}$ as the first step of Fig.~\ref{fig:pipeline}(a).
The function $f_\text{T}(\cdot)$ is to build the triangle-based face representation where the landmarks serve as vertices of the triangles.
The obtained face representations from $\mathbf{I_t}$ and $\mathbf{I}$ have the same number of triangles and those triangles correspond one by one according to the landmarks.
As we have the correspondence of triangles between the two face representations, $f_\text{R}(\cdot)$ transforms each triangle in $f_\text{T}(f_\text{L}(\mathbf{I}_\mathbf{t}))$ into the corresponding triangle in $f_\text{T}(f_\text{L}(\mathbf{I}))$ by affine transformation to get a full-face mask at the right side of Fig.~\ref{fig:pipeline}(a).
The function $f_\text{C}(\cdot)$ connects the landmarks of the contour of the lower face and the landmark of the nose (red dots in the full-face mask) in turn to obtain the faced mask area.
As shown in the beginning of Fig.~\ref{fig:pipeline}(b), after getting the facial mask $\mathbf{I_\text{M}}$, we overlay it on the original image $\mathbf{I}$ to get the Delaunay-based masked image $\mathbf{I_\text{DM}}\in \mathbb{R}^{H_1\times W_1\times 3}$ through
\begin{align}\label{eq:faceswap-2}
    \mathbf{I_\text{DM}} = f_2(\mathbf{I}, \mathbf{I_\text{M}}) = f_2(\mathbf{I},f_1(\mathbf{I}, \mathbf{I_t})).
\end{align}

To better motivate our proposed method, we have carried out a  pilot study.
Here, we briefly discuss the results using images obtained by DM in face recognition and mask detection and compare them with the results of solid-color masked faces.
We use the whole gallery set (1M images) and take 3,530 faces of 80 celebrities as the probe set from MegaFace Challenge 1 \cite{kemelmacher2016megaface}.
We compare the top-1 identification rates and the mask detection rates of images wearing solid-color medical masks and images generated by DM in Fig.~\ref{fig:bar}.
We can see that both indicators are very high when adding solid-colored medical masks, indicating that such masks hardly influence those discriminators.
Images obtained by DM have a considerable impact to face recognizers and the mask detector. 
However, DM is not effective enough as shown in the second cluster of Fig.~\ref{fig:bar}.
There are still about 51\% and 28\% tasks judged correctly for top-1 identification and mask detection, respectively, far from zero.
In order to strengthen the aggressiveness of the designed faced mask, we propose adversarial masking methods to add special textures, as explained in the following sections. 

The main challenges stem from:
\ding{182} Most of the face information is concentrated in the eye region.
In contrast, the part of the mask area, \eg, the mouth, plays a relatively low role in the face recognition task, which increases the difficulty of our work.
\ding{183} Although a part of images processed by DM can remain undetected by the facial mask detector, 28\% of images are detected due to the unavoidable factors in the process of adding masks (\eg, the chromatic aberration between faced masks and skins, discontinuities in textures), which makes interference to the mask detector unsuccessful.
\ding{184} Our goal is to combine multiple tasks, and it is difficult to simultaneously handle tasks that have different optimization directions.
\ding{185} Different deep-learning-based discriminators use different network structures and different network parameters.
It is hard to ensure the transferability that the generated faced masks can effectively interfere with diverse discriminators.

\subsection{Adversarial Noise Delaunay-based Masking}
\label{subsec:adversarial_noise_attack}
Inspired by adversarial attacks, \eg, project gradient descent (PGD) \cite{madry2017towards}, which performs effectively when targeting pre-trained neural networks, we apply an adversarial attack to the masked image $\mathbf{I_{DM}}$ acquired by DM.
We define this method as an adversarial noise Delaunay-based masking method (AdvNoise-DM) and show the process in the middle of Fig.~\ref{fig:pipeline}(b).
We replace $\mathbf{I_{DM}}$ with Eq.~\eqref{eq:faceswap-2} and generate the adversarial noise $\mathbf{n}\in \mathbb{R}^{H_1\times W_1\times 3}$ to obtain
\begin{align}
    \hat{\mathbf{I}} = f_2(\mathbf{I},f_1(\mathbf{I}, \mathbf{I_t})) + \mathbf{n},
    \label{eq:addnoise}
\end{align}
which denotes the superimposition of the intermediate $f_2(\mathbf{I},f_1(\mathbf{I}, \mathbf{I_t}))$ and the adversarial perturbation $\mathbf{n}$. 
%
%
Our goal is to find the $\hat{\mathbf{I}}$ which can not only mislead the FRS but also remain undetected by the mask detector by the means of obtaining such an adversarial perturbation $\mathbf{n}$.
For this reason, the problem to be solved can be transformed into achieving the optimal trade-off between face recognition and mask detection. Then, we have the following objective function
\begin{align}
    \argmax_{\mathbf{n}} \mathcal{D}(\text{FR}(f_2(\mathbf{I},f_1(\mathbf{I}, \mathbf{I_t})) + \mathbf{n}),\text{FR}(\mathbf{I})) \nonumber\\
    -\alpha * \mathcal{J}(\text{MD}(f_2(\mathbf{I},f_1(\mathbf{I}, \mathbf{I_t})) + \mathbf{n}),y).
    \label{eq:advnoise_obj}
\end{align}
In the first part of the objective function, $\text{FR}(\cdot)$ denotes a face recognition function which receives an image and returns the corresponding embedding.
$\mathcal{D}(\cdot)$ denotes the Euclidean distance between the embedding from the original image $\mathbf{I}$ and the embedding from the image $\hat{\mathbf{I}}$ processed by AdvNoise-DM. 
We intend to maximize this part of the objective function for the purpose of enlarging the gap of identification information before and after the modification. 

In the second part, $\text{MD}(\cdot)$ represents a mask detection function which receives an image and returns the probability of wearing a mask.
$\mathcal{J}(\cdot)$ denotes the cross-entropy loss function, y is the ground truth label for whether the face is masked, 0 for not masked, and 1 for masked.
Here, we set $y=0$ to force the image $\hat{\mathbf{I}}$ to be judged without a mask. 
The ratio $\alpha$ is used to adjust the focus between face recognition and mask detection. 
We aim to minimize this cross-entropy loss so we take a minus sign for this item.

The images obtained by AdvNoise-DM can significantly interfere with the discrimination of face recognizers and the mask detector.
This method can almost reduce the top-1 identification rate to zero and reduce the mask detection rate to only 7.3\% as shown in Fig.~\ref{fig:bar}.
Nevertheless, AdvNoise-DM has a certain drawback as it causes great changes to the pixels, so reduces the naturalness of the generated images.
In this case, it is necessary to use a smoother masking method.

\begin{figure}[tbp]
\centering
\includegraphics[width=1\columnwidth]{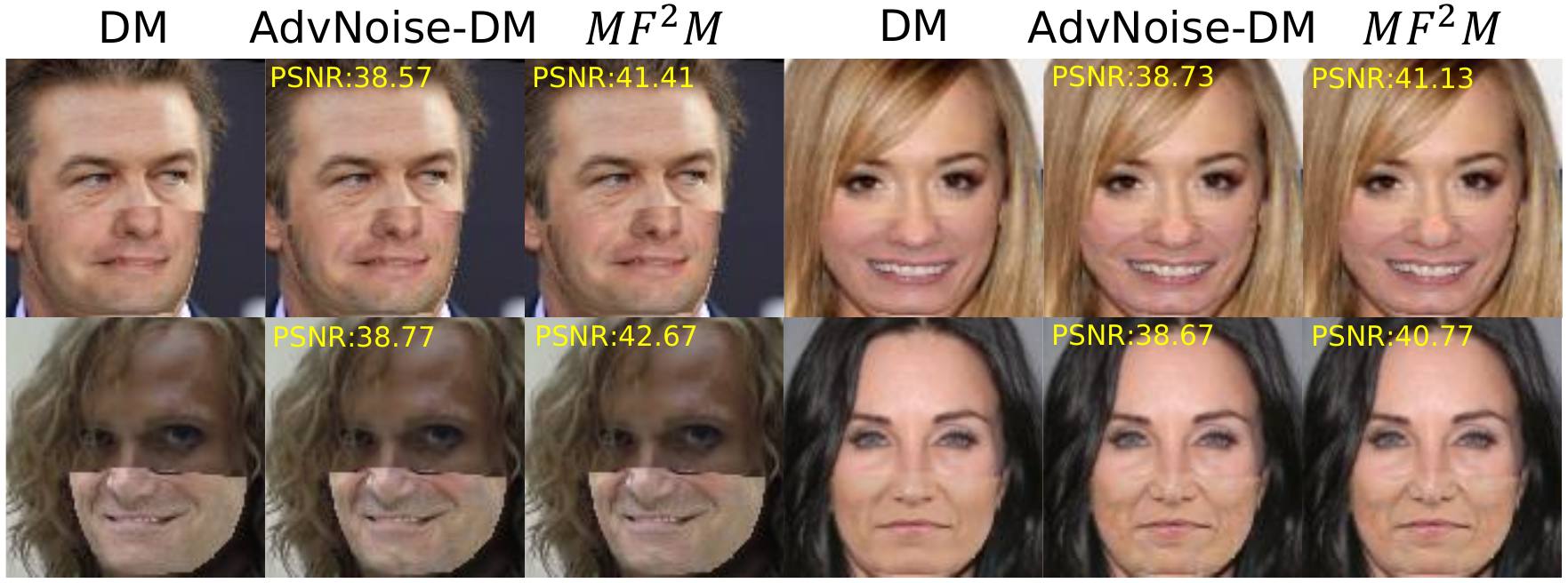}
\vspace{-20pt}
\caption{Examples to compare the PSNR values between AdvNoise-DM and $\text{MF}^2\text{M}$.
The PSNR value is calculated by the current image and the DM image.
}
\label{fig:psnr_compare}
\vspace{-10pt}
\end{figure}

\subsection{Adversarial Filtering Delaunay-based Masking}

\label{subsec:filtering-based_attack}
Since the filtering process brings better smoothness,
calculating each pixel by the surrounding pixels, we further propose an adversarial filtering Delaunay-based masking method ($\text{MF}^2\text{M}$).
This method combines noise-based and filtering-based attacks as shown in Fig.~\ref{fig:pipeline}(b).
We first apply DM and add a relatively small adversarial perturbation $\mathbf{n}$ to get the intermediate $f_2(\mathbf{I},f_1(\mathbf{I}, \mathbf{I_t})) + \mathbf{n}$, referring to the method AdvNoise-DM.
Then we utilize pixel-wise kernels $\mathbf{K}\in \mathbb{R}^{H_1\times W_1\times K^2}$ to process the intermediate. 
The $p$-th pixel of the intermediate $f_2(\mathbf{I},f_1(\mathbf{I}, \mathbf{I_t})) + \mathbf{n}$ is processed by the corresponding $p$-th kernel in $\mathbf{K}$, denoted as $\mathbf{K}_p\in \mathbb{R}^{K\times K}$, where $K$ represents the kernel size.
We retouch the original image $\mathbf{I}$ via the guidance of filtering and reformulate Eq.~\eqref{eq:addnoise} as
\begin{align}
    \tilde{\mathbf{I}} = \mathbf{K}\circledast{(f_2(\mathbf{I},f_1(\mathbf{I}, \mathbf{I_t})) + \mathbf{n})} 
    \label{eq:imagefilter},
\end{align}
where $\circledast$ denotes the pixel-wise filtering process and $\tilde{\mathbf{I}}\in \mathbb{R}^{H_1\times W_1\times 3}$ represents the filtered images.
In the $\text{MF}^2\text{M}$ procedure, we aim at obtaining a deceptive $\tilde{\mathbf{I}}$ in both face recognition task and mask detection task by altering the pixel-wise kernels $\mathbf{K}$. 
The objective function for optimization looks similar to Eq.~\eqref{eq:advnoise_obj} as following
\begin{align}
    \argmax_{\mathbf{K}} \mathcal{D}(\text{FR}(\mathbf{K}\circledast{(f_2(\mathbf{I},f_1(\mathbf{I}, \mathbf{I_t}))+\mathbf{n})}),\text{FR}(\mathbf{I})) \nonumber\\
    -\beta * \mathcal{J}(\text{MD}(\mathbf{K}\circledast{(f_2(\mathbf{I},f_1(\mathbf{I}, \mathbf{I_t}))+\mathbf{n}})),y).
    \label{eq:advfilter_obj}
\end{align}
Compared with Eq.~\eqref{eq:advnoise_obj}, the optimization objective becomes $\mathbf{K}$.
We intend to increase the Euclidean distance between the embedding from the original image $\mathbf{I}$ and that from the filtered image $\tilde{\mathbf{I}}$.
Meanwhile, we try to improve the probability that the filtered image $\tilde{\mathbf{I}}$ is judged not wearing a mask.
The ratio of the mask detection part is marked as $\beta$. 
As shown in Fig.~\ref{fig:bar}, almost all images generated by $\text{MF}^2\text{M}$ is deceptive for face recognition and only 6.95\% images are detected wearing masks.
Besides, $\text{MF}^2\text{M}$ brings higher naturalness than AdvNoise-DM.
Fig.~\ref{fig:psnr_compare} shows that the peak signal-to-noise ratio (PSNR) calculated between $\text{MF}^2\text{M}$ and DM is higher than that calculated between AdvNoise-DM and DM, indicating that $\text{MF}^2\text{M}$ changes images less.
We will use two similarity metrics in the experiment to prove this strong point.

\subsection{Algorithm for $\text{MF}^2\text{M}$}

Algorithm~\ref{alg:filtering_masking_algorithm} summarizes our method. 
First, we apply DM to complete the face replacing process, \ie, generating a faced mask extracted from $\mathbf{I_t}$ and overlaying it on $\mathbf{I}$ to obtain $\mathbf{I_{DM}}$.
Second, we add an adversarial noise $\mathbf{n}$ to $\mathbf{I_{DM}}$ and obtain $\hat{\mathbf{I}}$.
In the filtering attack process, we initialize the filtering kernels $\mathbf{K}$ whose initial action is to make the filtered image consistent with the original image (\ie, the weight of the center position of each kernel is 1, and the weight of other positions is 0).
In each iteration, we perform pixel-wise filtering by current kernels $\mathbf{K}$ and $\mathbf{I_{DM}}$ to acquire the current filtered image $\mathbf{I}^{'}$.
Then we calculate $\mathbf{Loss_D}$ and $\mathbf{Loss_{CE}}$, via the Euclidean distance function and the cross-entropy loss function, respectively.
These two loss functions constitute the final optimization objective function by the ratio $\beta$.
At the end of each iteration, we update the filtering kernels $\mathbf{K}$ according to the product of the step size $\epsilon$ and the gradient of the optimization objective.
Finally, we use the optimized kernels to embellish the aimed image $\tilde{\mathbf{I}}$.

\setlength{\textfloatsep}{0.1cm}
\begin{algorithm}[tb]
 	{
		\caption{$\text{MF}^2\text{M}$}
		\label{alg:filtering_masking_algorithm}
		\KwIn{Original image $\mathbf{I}$, Template image  $\mathbf{I_t}$, Ratio $\beta$,
		Face recognizer $\text{FR}(\cdot)$, Mask detector $\text{MD}(\cdot)$, Step size $\epsilon$,
		Label $y$ of not masked image, Iteration period $\mathbf{T}$.}
		\KwOut{Reconstruction image $\tilde{\mathbf{I}}$.}
		Generate $\mathbf{I_{DM}}$ by extracting the faced mask from $\mathbf{I_t}$ and overlay it on $\mathbf{I}$ by Delaunay triangulation.
		\\
		$\hat{\mathbf{I}} = \text{PGD}_{attack}(\mathbf{I_{DM}})$.
		\\
		Initial filtering kernels $\mathbf{K}$;
		\\
 		\For{$i=1\ \mathrm{to}\ \mathbf{T}$}{
 		    Generate filtered image $\mathbf{I}^{'}$ via
 		    $\mathbf{I}^{'} = \mathbf{K}\circledast{\hat{\mathbf{I}}}$;
 		    \\
 		    Calculate $\mathbf{Loss_D}$ via Euclidean distance function $\mathcal{D}$ $\mathbf{Loss_D}=\mathcal{D}(\text{FR}(\mathbf{I}^{'}),\text{FR}(\mathbf{I}))$;
 		    \\
 		    Calculate $\mathbf{Loss_{CE}}$ via cross-entropy loss function $\mathcal{J}$ $\mathbf{Loss_{CE}}=\mathcal{J}(\text{MD}(\mathbf{I}^{'}),y)$;
 		    \\
 		    Calculate the sum loss function $\mathbf{Loss}$ via 
 		    $\mathbf{Loss}=\mathbf{Loss_D} - \beta * \mathbf{Loss_{CE}}$;
 		    \\
 		    Update filtering kernels $\mathbf{K}$ via 
 		    $\mathbf{K}=\mathbf{K} + \epsilon * \mathop{\nabla_{\mathbf{K}}}\mathbf{Loss}$;
    	}
    	 Apply image filtering to obtain reconstruction image $\tilde{\mathbf{I}}$ via $\tilde{\mathbf{I}} = \mathbf{K}\circledast \hat{\mathbf{I}}$\;
	}
	
\end{algorithm}
\setlength{\floatsep}{0.1cm}


\section{Experiments}

\subsection{Experimental Setup}
\label{sec:setting}

\noindent\textbf{Face recognition methods.}
In our white-box attack experiment, the backbone of the face recognizer \cite{arcface_torch} is pre-trained ResNet50 under ArcFace. 
The face recognizer takes cropped images (112 $\times$ 112) as input and returns the final 512-D embedding features. 
To illustrate  the transferability, we further use recognizers \cite{cosface_torch} pre-trained under CosFace with ResNet34 and ResNet50 as the backbone respectively to verify the black-box attack performance.
We choose these two FRS as ArcFace and CosFace perform SOTA in face recognition.

\vspace{5pt}
\noindent\textbf{Mask detection methods.}
The mask detection method bases on RetinaNet \cite{lin2017focal}, an efficient one-stage objects detecting method.
The pre-trained model \cite{mask_detection} we used is competitive in existing mask detectors, achieving 91.3\% mAP at the face\_mask validation dataset (including 1839 images).
The mask detector outputs two probabilities of not-masked and masked faces respectively.
By comparing these two probabilities, we can judge whether there is a mask.

\vspace{5pt}
\noindent\textbf{Datasets.}
We utilize 1M images of 690K individuals in MegaFace Challenge 1 as the gallery set. 
In terms of the probe set, we refer to the setting of MegaFace \cite{kemelmacher2016megaface} and use a subset of FaceScrub (\ie, 3,530 images of 80 celebrities) for efficiency.
For the template images used to extract faced masks, we use StyleGAN to generate images with seeds numbered from 1 to 13,000.
As some generated images have illumination or occlusion problems, we manually select 3,136 high-quality face images.

\vspace{5pt}
\noindent\textbf{Evaluation settings.} 
The face recognition evaluation is based on masked/not-masked pairs.
We add masks to images of the probe set and remain images in the gallery not-masked.
When adding masks, we select the most similar face image to the original face image from 3,136 template images according to the features extracted by the face recognition model, which aims to make the masked faces look more natural.
In AdvNoise-DM, we use PGD attack to add deliberate noise.
The epsilon (maximum distortion of adversarial example) is 0.04. The step size for each attack iteration is 0.001 while the number of iterations is 40.
The ratio $\alpha$ is set to 1.
In $\text{MF}^2\text{M}$, we add noise with an epsilon of 0.01.
The kernel size of the pixel-wise kernels is 5.
When alter the pixel-wise kernels, the step size is 0.1 and the number of iterations is 160.
Here we set the coefficient $\beta$ to 1, same as $\alpha$.
All optimization objectives are restricted to the faced mask area obtained by a deep learning-based method \cite{coordinate}.

\vspace{5pt}
\noindent\textbf{Baseline methods.}
We apply seven SOTA adversarial attack methods to the solid-colored medical masks as our baselines, including FGSM, I-FGSM, MI-FGSM, TI-FGSM, TI-MI-FGSM, $\text{DI}^2$-FGSM, and M-$\text{DI}^2$-FGSM.
The epsilon of these baselines is 0.04, to maintain the same attack intensity as AdvNoise-DM.
The iterations number involved in baselines is 40, and the momentum coefficient involved is 1.
The added perturbation is restricted to the facial mask area.

\vspace{5pt}
\noindent\textbf{Metrics.}
For face recognition, we use the top-1 identification rate in the face identification task, the true accept rate (TAR) at $10^{-6}$ false accept rate (FAR) and the area under curve (AUC) in the face verification task. 
For mask detection, we use the detection rate.
To reflect the degree of reconstructed modification and evaluate the naturalness of the generated images, we further use the PSNR and the structural similarity (SSIM) \cite{wang2004image} to measure the similarity between the adversarial masked results and the corresponding images from DM.
The region of the calculation for similarity metrics is the whole image. 

\subsection{Comparison on White-box Attack}
\label{sec:white_box}

Face recognition has two main tasks, face identification and face verification.
Given a probe image and a gallery, identification aims to find an image which has the same identity as the probe image from the gallery, \ie, 1 vs. $N$ search task.
The verification task sets a threshold to judge whether two images have the same identity, \ie, 1 vs. 1 comparison task.
%


\begin{figure*}
\centering 
\includegraphics[width=1.8\columnwidth]{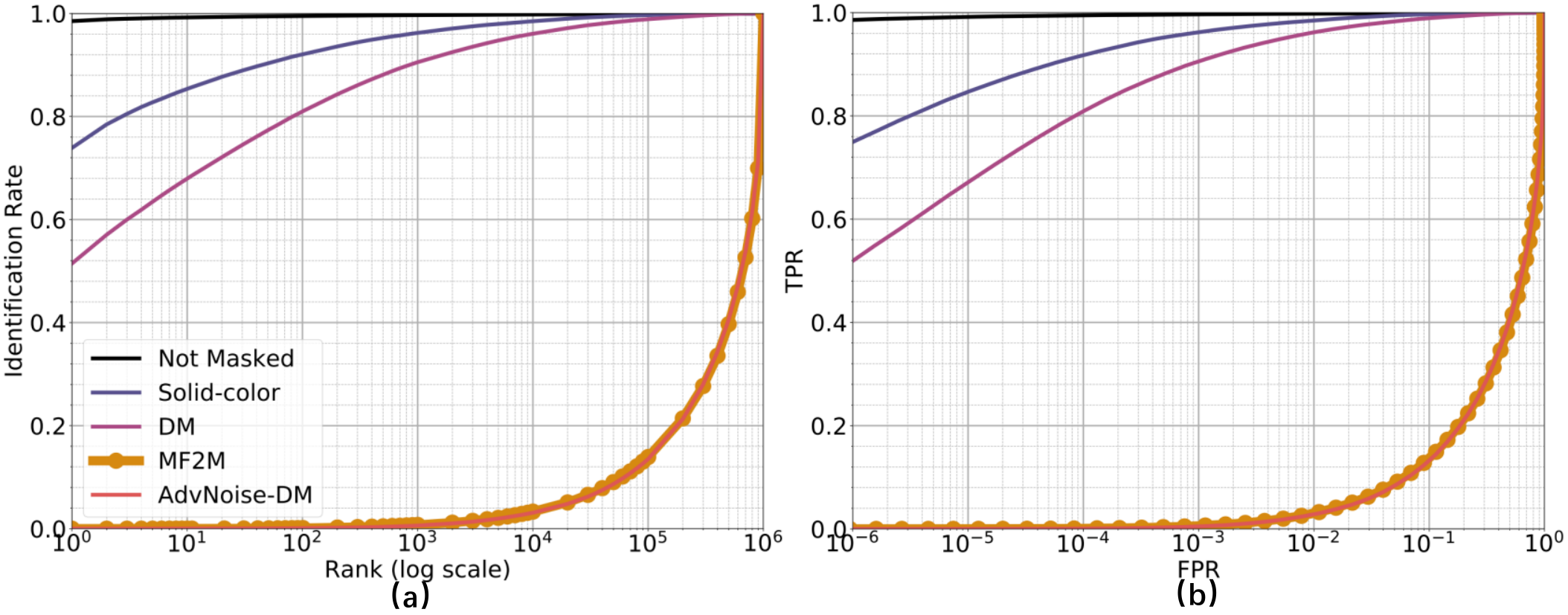}
\vspace{-10pt}
\caption{The effects of different masking methods on face recognition in white-box attacks. 
(a) The CMC curves of the face identification task. The abscissa indicates the number of images selected from the gallery. The ordinate denotes the identification rate at the specified number of images.
(b) The ROC curves of the face verification task. 
The abscissa indicates the FPR and the ordinate denotes the TPR.
}
\label{fig:white_box_result}
\end{figure*}
\begin{table}[t]
    \centering
    \caption{Multifaceted evaluation of each masking method on white-box attack. 
"Rank 1" refers to the top-1 identification rate and "Veri." refers to the TAR at $10^{-6}$ FAR.
}
    \vspace{-10pt}
    \resizebox{1\linewidth}{!}{
    \begin{tabular}{l|cccc}
\hline 
 & \multicolumn{1}{c}{Face Rec.} & \multicolumn{2}{c}{\cellcolor{gray!40}Face Verification} & \multicolumn{1}{c}{Mask Detection}\tabularnewline
\cline{2-5} \cline{3-5} \cline{4-5} \cline{5-5}
 & \multicolumn{1}{c}{Rank 1} & \multicolumn{1}{c}{Veri.} & \multicolumn{1}{c}{AUC} & \multicolumn{1}{c}{Mask Rate} \tabularnewline
\hline 
Solid-color & 0.7389 & 0.7470 & 0.9986 & 90.53\%  \tabularnewline
\hline 
FGSM & 0.4895 & 0.4924 & 0.9950 & 62.10\%  \tabularnewline
I-FGSM & 0.0039 & 0.0026 & 0.7594 & 39.41\%  \tabularnewline
MI-FGSM & 0.0179 & 0.0135 & 0.8667 & 43.91\%  \tabularnewline
TI-FGSM & 0.4991 & 0.5053 & 0.9950 & 66.00\%  \tabularnewline
TI-MI-FGSM & 0.0364 & 0.0290 & 0.8998 & 51.97\%  \tabularnewline
$\text{DI}^2$-FGSM & 0.0690 & 0.0601 & 0.9193 & 43.51\%  \tabularnewline
M-$\text{DI}^2$-FGSM & 0.0705 & 0.0623 & 0.9293 & 45.50\% \tabularnewline
\hline
\textbf{DM (ours)} & 0.5146 & 0.5154 & 0.9956 & 27.74\%  \tabularnewline
\textbf{AdvNoise-DM (ours)} & \textbf{1.35$\text{e}^{-5}$} & \textbf{1.35$\text{e}^{-5}$} &  0.4163 & 7.30\% \tabularnewline
\textbf{$\text{MF}^2\text{M}$ (ours)} & \textbf{1.35$\text{e}^{-5}$} & 2.02$\text{e}^{-5}$ & \textbf{0.4093} & \textbf{6.95\%} \tabularnewline
\hline


\end{tabular}
}

\label{Table:table_white}

\vspace{5pt}

\end{table}

\vspace{5pt}
\noindent\textbf{Face identification.}
We constitute 151K pairs with the same identity from 3,530 face images of 80 celebrities.
For each pair, we take one image as the probe image and put the other image into the gallery.
Top-k identification rate denotes the successful rate of matching pairs where k is the number of images selected from the gallery.
Fig.~\ref{fig:white_box_result}(a) shows the cumulative matching characteristic (CMC) curves of images under different masking states.
The abscissa indicates the number of images selected from the gallery according to the embedding obtained by the face recognizer.
The ordinate denotes the identification rate at the specified number of images.
When images are without masks, the top-1 identification rate (\ie, ``Rank 1'') is 0.98, proving that the recognizer achieves good performance without face occlusion.
After adding solid-color medical masks, ``Rank 1'' reduces to 0.7389.
This metric for DM declines to 0.5146.
As for AdvNoise-DM and $\text{MF}^2\text{M}$, 
the higher the attack intensity, the more their corresponding curves are close to the lower right of the graph.
We respectively  alter the iterations numbers and make the performance of AdvNoise-DM and $\text{MF}^2\text{M}$ close in this task, so as to compare them on other indicators.
``Rank 1'' of both methods drop to \textbf{1.35$\text{e}^{-5}$}, indicating that the SOTA face recognizer performs poorly under AdvNoise-DM and $\text{MF}^2\text{M}$.
The second column of Table~\ref{Table:table_white} shows that AdvNoise-DM and $\text{MF}^2\text{M}$ achieve significantly lower ``Rank 1'' than seven SOTA additive-perturbation-based baselines.
%
%

\vspace{5pt}
\noindent\textbf{Face verification.}
We use the 3,530 images of 80 identities in the probe set and 1M images in the gallery to build 151K positive samples and 3.5 billion negative samples for face verification.
Fig.~\ref{fig:white_box_result}(b) shows the receiver operating characteristic (ROC) curves.
We define the true positive rate (TPR) when the false positive rate (FPR) is $1e^{-6}$ as ``Veri.'', which is 0.7470 and 0.5154
for solid-color medical masks and DM, respectively.
When we apply AdvNoise-DM and $\text{MF}^2\text{M}$, ``Veri.'' almost drops to zero.
In the third column and the fourth column of Table~\ref{Table:table_white}, we show the ``Veri.'' values and the AUC values, respectively.
We can see that both metrics of AdvNoise-DM and $\text{MF}^2\text{M}$ are much lower than baselines.

\vspace{5pt}
\noindent\textbf{Mask detection.}
We exhibit the mask detection rate of different masking methods in the fifth column of Table~\ref{Table:table_white}, which also represents the accuracy of the used mask detector.
Solid-colored masks are easily detected and the detection rate is 90.53\%.
DM reduces the detection rate to 27.74\%.
AdvNoise-DM further interferes with the judgment of the detector and the accuracy decreases to only 7.30\%.
$\text{MF}^2\text{M}$ achieves the best attack performance and reduces this rate to 6.95\%.
The detection rates for additive-perturbation-based baselines are between 39\% and 66\%.
So far, we prove that our adversarial methods are very effective for both face recognition and mask detection in white-box attacks.

\vspace{5pt}
\noindent\textbf{Similarity measurement.}
Now we turn to the discussion upon the similarity measurement before and after adding adversarial textures.
The value of similarity is calculated by comparing with images obtained by DM, so we only calculate similarity scores for AdvNoise-DM and $\text{MF}^2\text{M}$.
We choose SSIM and PSNR as our similarity metrics, evaluating the similarity from aspects of visual error and structure difference.
The SSIM of AdvNoise-DM and $\text{MF}^2\text{M}$ are 0.9808 and 0.9812, respectively, indicating the high structural similarity in the reconstruction process.
In terms of PSNR, the value of $\text{MF}^2\text{M}$ is 40.45, higher than 38.76 of AdvNoise-DM, which is in line with the cognitive experience that the filtering operation has a more imperceptible modification to images.

\subsection{Comparison on Black-box Attack Transferability}
\label{sec:black_box}

\begin{figure*}[t]
	\centering 
	\includegraphics[width=1.8\columnwidth]{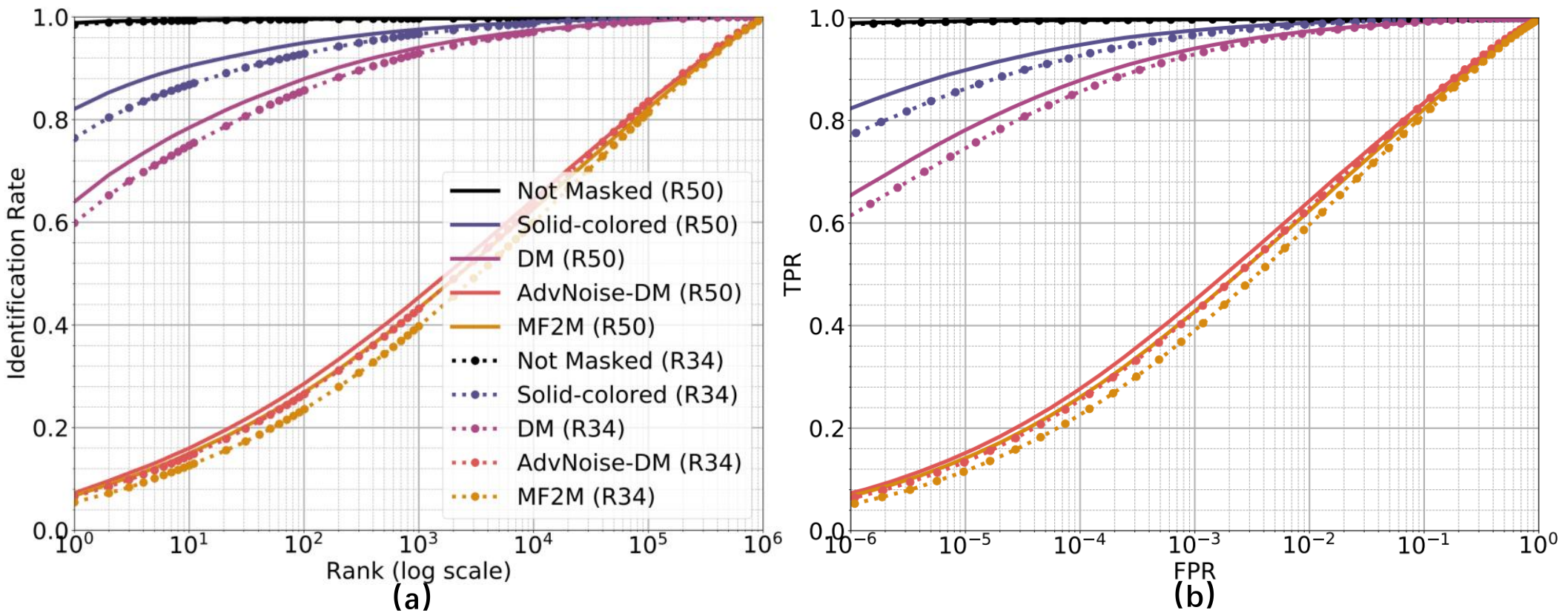}
	\vspace{-10pt}

    \caption{The effects of different masking methods on face recognition in black-box attacks. 
    (a) The CMC curves of the face identification task. 
    (b) The ROC curves of the face verification task.
    The abscissa and ordinate are consistent with Fig.~\ref{fig:white_box_result}.
    ~R34 and R50 in the legend indicate that ResNet34 and ResNet50 are separately used as the backbone.
    }
    \label{fig:black_box_result}
\end{figure*}
%
\begin{table}[t]
    \centering
    \caption{Multifaceted evaluation of each masking method on black-box attack.
``Rank 1'' refers to the top-1 identification rate and ``Veri.'' refers to the TAR at $10^{-6}$ FAR.
}
    \vspace{-10pt}
    \resizebox{1\linewidth}{!}{
   \begin{tabular}{l|ccc|ccc}
\hline 
 & \multicolumn{3}{c|}{CosFace ResNet50} & \multicolumn{3}{c}{CosFace ResNet34}\tabularnewline
\cline{2-7} \cline{3-7} \cline{4-7} \cline{5-7} \cline{6-7} \cline{7-7} 
 & Face Rec. & \multicolumn{2}{c|}{\cellcolor{gray!40}Face Verification} & Face Rec. & \multicolumn{2}{c}{\cellcolor{gray!40}Face Verification}\tabularnewline
\cline{2-7} \cline{3-7} \cline{4-7} \cline{5-7} \cline{6-7} \cline{7-7} 
 & Rank 1 & Veri. & AUC & Rank 1 & Veri. & AUC\tabularnewline
\hline 
Solid-color & 0.8207 & 0.8233 & 0.9991 & 0.7641 & 0.7697 & 0.9987\tabularnewline
\hline 
FGSM & 0.7585 & 0.7624 & 0.9988 & 0.7147 & 0.7212 & 0.9984\tabularnewline
I-FGSM & 0.6720 & 0.6752 & 0.9978 & 0.6243 & 0.6303 & 0.9967\tabularnewline
MI-FGSM & 0.6419 & 0.6445 & 0.9972 & 0.6020 & 0.6077 & 0.9961\tabularnewline
TI-FGSM & 0.7640 & 0.7711 & 0.9988 & 0.7142 & 0.7214 & 0.9984\tabularnewline
TI-MI-FGSM & 0.6416 & 0.6438 & 0.9972 & 0.5996 & 0.6038 & 0.9958\tabularnewline
$\text{DI}^2$-FGSM & 0.7178 & 0.7222 & 0.9983 & 0.6699 & 0.6777 & 0.9977\tabularnewline
M-$\text{DI}^2$-FGSM & 0.6559 & 0.6592 & 0.9974 & 0.6165 & 0.6242 & 0.9963\tabularnewline
\hline 
\textbf{DM (ours)} & 0.6400 & 0.6520 & 0.9971 & 0.5987 & 0.6126 & 0.9971\tabularnewline
\textbf{AdvNoise-DM (ours)} & 0.0726 & 0.0723 & 0.9293 & 0.0657 & 0.0622 & 0.9310\tabularnewline
\textbf{$\text{MF}^2\text{M}$ (ours)} & \textbf{0.0674} & \textbf{0.0670} & \textbf{0.9237} & \textbf{0.0546} & \textbf{0.0508} & \textbf{0.9212}\tabularnewline
\hline 
\end{tabular}
}

\label{Table:table_black}

\vspace{5pt}

\end{table}
To verify the transferability of our methods, we utilize generated masked images to carry out black-box attacks.
We conduct black-box attacks against face recognition models pre-trained under Cosface with ResNet34 and ResNet50 as the backbone.
Curves with dots and without dots in Fig.~\ref{fig:black_box_result}  represent results of attacking ResNet34 and ResNet50, respectively.
Compared with the ``Rank 1'' of AdvNoise-DM and $\text{MF}^2\text{M}$ in Fig.~\ref{fig:white_box_result}(a) (nearly zero), they vary from 0.05 to 0.08 in Fig.~\ref{fig:black_box_result}(a). 
It shows that the interference degree to face recognizers in black-box attacks reduces, but is still strong, \ie, our adversarial masking methods have sufficient transferability.
Besides, curves of $\text{MF}^2\text{M}$ are lower than curves of AdvNoise-DM targeting the same model, indicating $\text{MF}^2\text{M}$ has stronger transferability.
Compared with adversarial attack baselines, AdvNoise-DM and $\text{MF}^2\text{M}$ have absolute advantages in the impact on face recognizers as shown in Table~\ref{Table:table_black}.

\subsection{Extension to Physical Attack}
\begin{figure}
\centering 
\includegraphics[width=1\columnwidth]{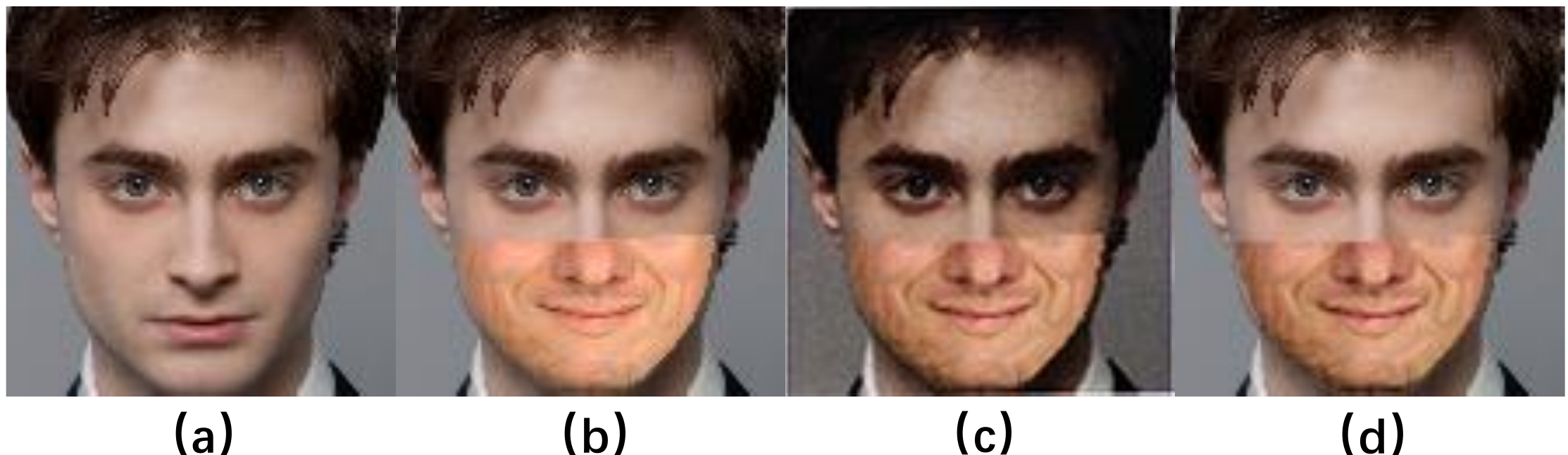}
\vspace{-15pt}
\caption{The process of synthesizing the image in the physical attack. (a) The original image in the FaceScrub dataset. (b) The digital image by $\text{MF}^2\text{M}$. (c) The recaptured image. (d) Extract the faced mask from (c) and overlay it to (a), used in the physical attack.}
\label{fig:scan}
\vspace{10pt}
\end{figure}

Due to various COVID-19 related restrictions, we were not able to recruit human subjects to study the effect of physical attack by wearing our proposed faced masks.
Therefore, we use an alternative recapture method to illustrate the physical effects of our proposed masking methods.
We randomly select 20 faces of different identities from the FaceScrub dataset as origin images in Fig.~\ref{fig:scan}(a).
We process the 20 faces with our proposed $\text{MF}^2\text{M}$ and obtain digital attacked faces as shown in Fig.~\ref{fig:scan}(b).
Then we use an $ApeosPort-IV C5575$ printer to print these attacked images and recapture images like Fig.~\ref{fig:scan}(c), which has obvious color differences from Fig.~\ref{fig:scan}(b). 
This procedure is meant for methodologically mimicking the process of plastering the patterns from a digital medium onto a physical one, such as fabric, linen, or paper, so that the physical appearance can be digitally reacquired via image sensors.
Finally, we resize the recaptured images to the size of 112 $\times$ 112, extract faced masks from them, overlay faced masks to the 20 corresponding original faces, and obtain images used in the physical attack as shown in Fig.~\ref{fig:scan}(d).
Based on this synthesis process, we conduct the experiment of physical attacks and demonstrate the robustness of our proposed $\text{MF}^2\text{M}$ and AdvNoise-DM to evade face recognition and mask detection.
Except for the solid-color medical mask and DM, we choose \textit{I-FGSM baseline} (\ie, apply I-FGSM to solid-color masks), which is the strongest baseline in digital attacks, as the main baseline for the physical attacks.
We also process these masking methods by the method shown in Fig.~\ref{fig:scan} and set three baselines in physical attacks.

\begin{figure}
\centering 
\includegraphics[width=1\columnwidth]{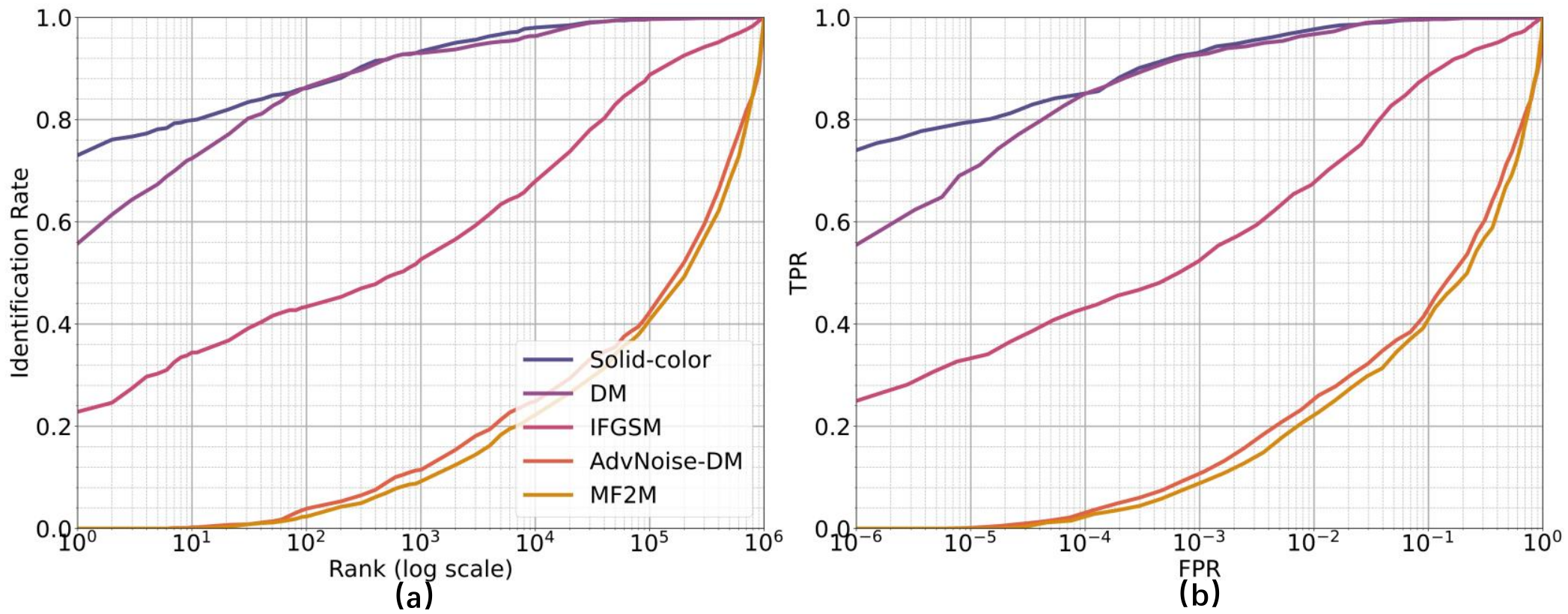}
\vspace{-10pt}
\caption{The effects of different masking methods on face recognition in \textit{physical} white-box attacks. 
(a) The CMC curves of the face identification task. 
(b) The ROC curves of the face verification task.
The abscissa and ordinate are consistent with Fig.~\ref{fig:white_box_result}.
}
\label{fig:white_physical}
\vspace{10pt}
\end{figure}

For the physical white-box attack, the second column and the third column of Table~\ref{Table:physical_white} show that the top-1 identification rates and the TAR at $10^{-6}$ FAR of $\text{MF}^2\text{M}$ and AdvNoise-DM are both zero, indicating the strong interference of these two masking methods to the face recognizer.
The last column of Table~\ref{Table:physical_white} shows that only 10\% of images by $\text{MF}^2\text{M}$ and AdvNoise-DM are detected faced masks, demonstrating the powerful ability of our proposed masking methods in avoiding mask detection in physical white-box attack.
The corresponding CMC curves in Fig.~\ref{fig:white_physical}(a) illustrate that 
the identification rates of $\text{MF}^2\text{M}$ and AdvNoise-DM are always below three baselines at different ranks.
The ROC curves in Fig.~\ref{fig:white_physical}(b) show that the TPR of $\text{MF}^2\text{M}$ and AdvNoise-DM are always less than three baselines at different FPR.

\begin{figure}
\centering 
\includegraphics[width=1\columnwidth]{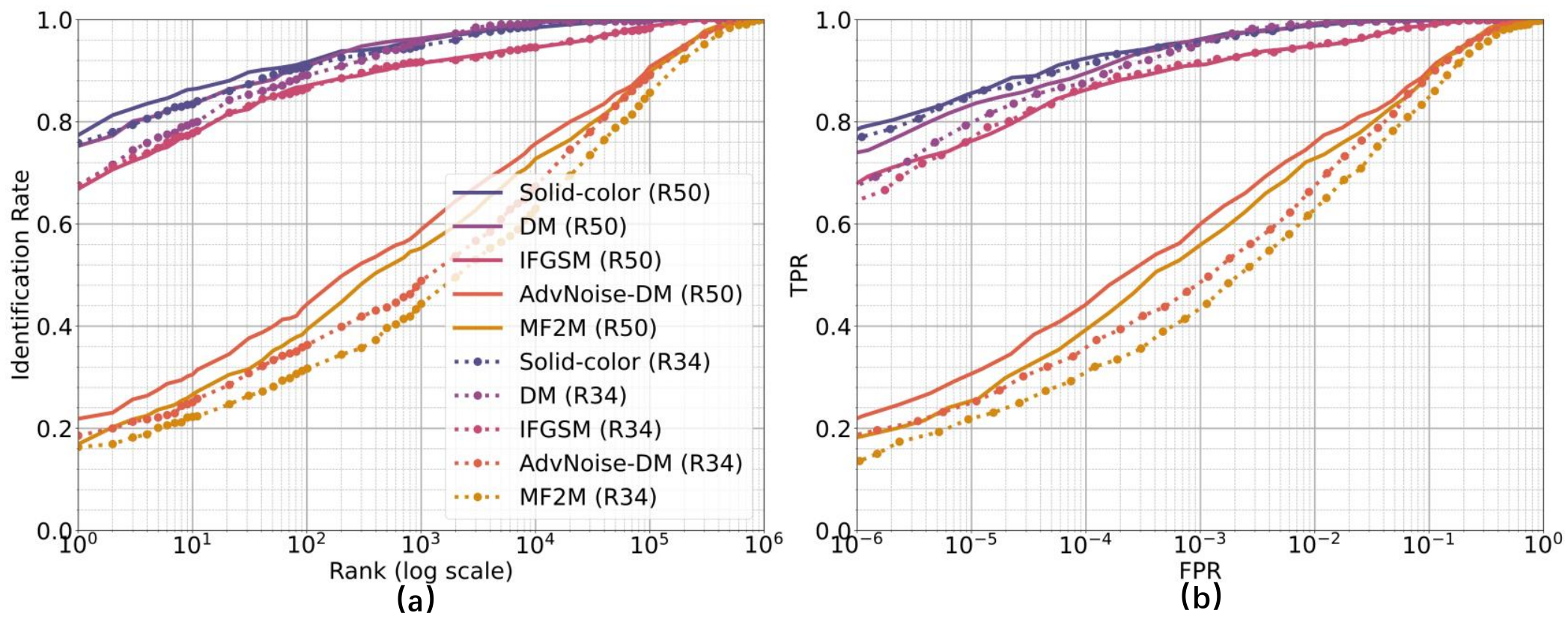}
\vspace{-10pt}

\caption{The effects of different masking methods on face recognition in \textit{physical} black-box attacks. 
(a) The CMC curves of the face identification task. 
(b) The ROC curves of the face verification task.
The abscissa and ordinate are consistent with Fig.~\ref{fig:white_box_result}.
R34 and R50 in the legend indicate that ResNet34 and ResNet50 are separately used as the backbone.}

\label{fig:black_physical}
\vspace{10pt}
\end{figure}

As for the physical black-box attack, the top-1 identification rates and the TAR at $10^{-6}$ FAR of \textit{I-FGSM baseline} are more than 0.64 as shown in Table~\ref{Table:physical_black}.
It indicates that the adversarial textures added in \textit{I-FGSM baseline} almost failed.
In contrast, these metrics of our proposed $\text{MF}^2\text{M}$ and AdvNoise-DM are still less than 0.22, indicating our proposed masking methods remain highly interference to face recognizers in physical black-box attack.
Compared with AdvNoise-DM, $\text{MF}^2\text{M}$ has a greater influence on face recognizers.
The corresponding CMC curves and ROC curves of physical black-box attack are shown in Fig.~\ref{fig:black_physical}.


\begin{table}[t]
    \centering
    \caption{Multifaceted evaluation of \textit{physical} white-box attacks. 
"Rank 1" refers to the top-1 identification rate and "Veri." refers to the TAR at $10^{-6}$ FAR.
}
    \vspace{-10pt}
    \resizebox{1\linewidth}{!}{
\begin{tabular}{l|cccc}
\hline 
 & \multicolumn{1}{c}{Face Rec.} & \multicolumn{2}{c}{\cellcolor{gray!40}Face Verification} & \multicolumn{1}{c}{Mask Detection}\tabularnewline
\hline 
 & \multicolumn{1}{c}{Rank 1} & \multicolumn{1}{c}{Veri.} & \multicolumn{1}{c}{AUC} & \multicolumn{1}{c}{Mask Rate}\tabularnewline
\hline 
Solid-color (phys.)& 0.7302 & 0.7398 & 0.9985 & 100\%\tabularnewline
DM (phys.)& 0.5574 & 0.5550 & 0.9975 & 10\%\tabularnewline
I-FGSM (phys.)& 0.2284 & 0.2497 & 0.9453 & 95\%\tabularnewline
\hline 
\textbf{AdvNoise-DM (phys.)} & \textbf{0.0000} & \textbf{0.0000} & 0.6879 & \textbf{10\%}\tabularnewline
\textbf{$\text{MF}^{2}\text{M}$ (phys.)} & \textbf{0.0000} & \textbf{0.0000} & \textbf{0.6658} & \textbf{10\%}\tabularnewline
\hline 
\end{tabular}
}
\vspace{10pt}

\label{Table:physical_white}

\vspace{5pt}

\end{table}
\begin{table}[t]
    \centering
    \caption{Multifaceted evaluation of \textit{physical} black-box attacks. 
"Rank 1" refers to the top-1 identification rate and "Veri." refers to the TAR at $10^{-6}$ FAR.
}
    \vspace{-10pt}
    \resizebox{1\linewidth}{!}{
\begin{tabular}{l|ccc|ccc}
\hline 
 & \multicolumn{3}{c|}{CosFace ResNet50} & \multicolumn{3}{c}{CosFace ResNet34}\tabularnewline
\cline{2-7} \cline{3-7} \cline{4-7} \cline{5-7} \cline{6-7} \cline{7-7} 
 & Face Rec. & \multicolumn{2}{c|}{\cellcolor{gray!40}Face Verification} & Face Rec. & \multicolumn{2}{c}{\cellcolor{gray!40}Face Verification}\tabularnewline
\cline{2-7} \cline{3-7} \cline{4-7} \cline{5-7} \cline{6-7} \cline{7-7} 
 & \multicolumn{1}{c}{Rank 1} & \multicolumn{1}{c}{Veri.} & AUC & Rank 1 & Veri. & AUC\tabularnewline
\hline 
Solid-color (phys.) & 0.7740 & 0.7846 & 0.9992 & 0.7586 & 0.7670 & 0.9993\tabularnewline
DM (phys.) & 0.7527 & 0.7369 & 0.9987 & 0.6757 & 0.6739 & 0.9985\tabularnewline
I-FGSM (phys.) & 0.6686 & 0.6791 & 0.9958 & 0.6734 & 0.6438 & 0.9951\tabularnewline
\hline 
\textbf{AdvNoise-DM (phys.)} & 0.2189 & 0.2184 & 0.9683 & 0.1858 & 0.1860 & 0.9627\tabularnewline
\textbf{$\text{MF}^{2}\text{M}$ (phys.)} & \textbf{0.1692} & \textbf{0.1818} & \textbf{0.9670} & \textbf{0.1633} & \textbf{0.1336} & \textbf{0.9498}\tabularnewline
\hline 
\end{tabular}
}

\label{Table:physical_black}

\vspace{10pt}

\end{table}

\subsection{Ablation Study and Discussion}
\label{sec:discussion}
\begin{figure}
\centering 
	\includegraphics[width=0.99\columnwidth]{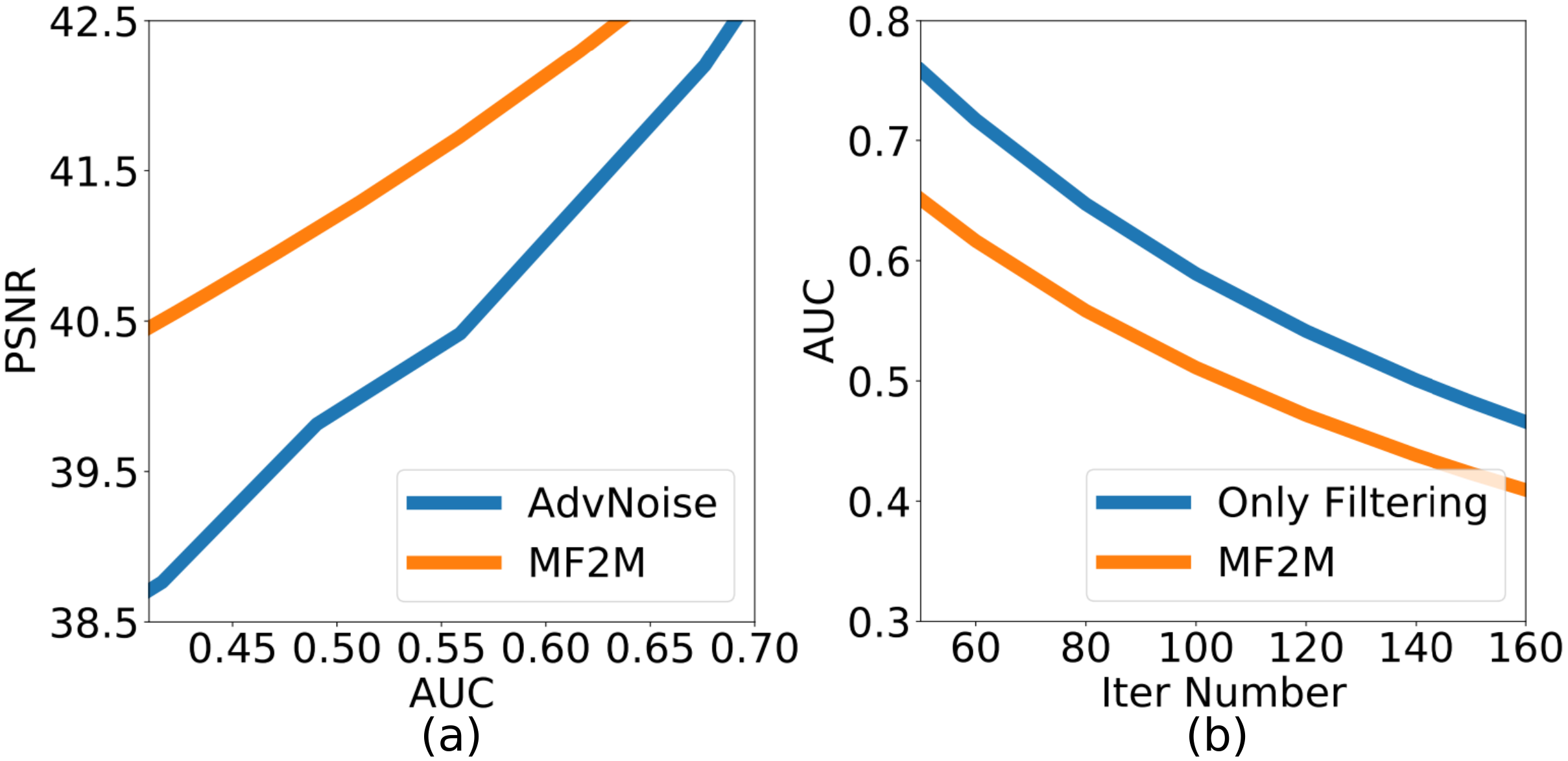}
	\vspace{-10pt}

\caption{(a) PSNR at different AUC of AdvNoise-DM and $\text{MF}^2\text{M}$.
(b) AUC at different iteration numbers of $\text{MF}^2\text{M}$ and the method only using filtering kernels.
}
\label{fig:ablation}
\vspace{10pt}
\end{figure}

%
In AdvNoise-DM, we only do noise-based adversarial attacks, and the intensity of noise reaches 0.04.
When we combine noise-based and filtering-based attacks in $\text{MF}^2\text{M}$, we only need lower noise intensity (\ie, 0.01) to achieve similar results in face recognition.
We experiment under different attack intensities and compare these two methods in the PSNR-AUC curves.
As shown in Fig.~\ref{fig:ablation}(a), $\text{MF}^2\text{M}$ (orange curve) gets higher PSNR values than AdvNoise-DM (blue curve) at the same AUC values, indicating the advantage of $\text{MF}^2\text{M}$ in naturalness.
We also conduct experiments on attacking images by only utilizing filtering kernels and show the comparison in Fig.~\ref{fig:ablation}(b).
At the same number of iterations for altering kernels, $\text{MF}^2\text{M}$ (orange curve) achieves lower AUC values than the method only using filtering kernels (blue curve).
This shows that adding noise can assist the optimization of filtering kernels as noise-based attacks have fewer parameters and higher attack efficiency.


\section{Conclusions}
In this paper, we propose $\text{MF}^2\text{M}$, an adversarial masking framework that adds faced masks containing partial face patterns and special adversarial textures.
Our work reveals the potential risks of existing face recognizers and mask detectors regarding facial masks specially customized.
The reconstructed images from our methods retain enough naturalness, generating a higher safety hazard.
Therefore, particularly generated facial masks should be taken into consideration when designing the FRS and mask detection systems.

\clearpage
\newpage
\bibliographystyle{ACM-Reference-Format}
\balance
\bibliography{ref}

\end{document}